 \documentclass[pmlr,twocolumn,10pt]{jmlr} % W&CP article

\usepackage{booktabs}
\usepackage[normalem]{ulem}
\useunder{\uline}{\ul}{}

\usepackage{booktabs}
\usepackage{siunitx}
\usepackage{dsfont}
\usepackage{xcolor}
\newcommand{\ours}{\textit{AdaMedGraph}}

\theorembodyfont{\upshape}
\theoremheaderfont{\scshape}
\theorempostheader{:}
\theoremsep{\newline}

\jmlrvolume{LEAVE UNSET}
\jmlryear{2023}
\jmlrsubmitted{LEAVE UNSET}
\jmlrpublished{LEAVE UNSET}
\jmlrworkshop{Machine Learning for Health (ML4H) 2023} % W&CP title

\title[AdaMedGraph]{AdaMedGraph: Adaboosting Graph Neural Networks for Personalized Medicine}

\author{%
\Name{Jie Lian}\thanks{Work done during an internship in Microsoft Research Asia.}\Email{jlian@connect.hku.hk}\\
\addr The University of Hong Kong, Hong Kong SAR, China
\AND
\Name{Xufang Luo}\thanks{Corresponding author.}\Email{xufluo@microsoft.com}\\
\addr Microsoft Research, China
\AND
\Name{Caihua Shan} \Email{caihua.shan@microsoft.com}\\
\addr Microsoft Research, China
\AND
\Name{Dongqi Han} \Email{dongqihan@microsoft.com}\\
\addr Microsoft Research, China
\AND
\Name{Varut Vardhanabhuti} \Email{varv@hku.hk}\\
\addr The University of Hong Kong, Hong Kong SAR, China
\AND
\Name{Dongsheng Li} \Email{dongsli@microsoft.com}\\
\addr Microsoft Research, China
}
\begin{document}
\maketitle

\begin{abstract}
Precision medicine tailored to individual patients has gained significant attention in recent times. Machine learning techniques are now employed to process personalized data from various sources, including images, genetics, and assessments. These techniques have demonstrated good outcomes in many clinical prediction tasks. Notably, the approach of constructing graphs by linking similar patients and then applying graph neural networks (GNNs) stands out, because related information from analogous patients are aggregated and considered for prediction. 
However, selecting the appropriate edge feature to define patient similarity and construct the graph is challenging, given that each patient is depicted by high-dimensional features from diverse sources. Previous studies rely on human expertise to select the edge feature, which is neither scalable nor efficient in pinpointing crucial edge features for complex diseases.
In this paper, we propose a novel algorithm named \ours, which can automatically select important features to construct multiple patient similarity graphs, and train GNNs based on these graphs as weak learners in adaptive boosting. \ours{} is evaluated on two real-world medical scenarios and shows superiors performance.

% Recent releases and publicly available healthcare data collection have yielded extensive datasets encompassing imaging, genetics, clinical assessments, and so on. Leveraging those multi-modality data for accurate disease prediction is important. In this context, graph-based methods have gained prominence by modeling relationships among patients, allowing us to represent individuals as nodes and capture pairwise similarities through edges. However, existing approaches often rely on manual edge selection and focus on a single relationship type.

% In response, we propose an automated graph construction method that considers multiple relations, offering transparency and interpretability for clinical insights. We evaluated our approach in two medical scenarios, and our AdaMedGraph showed superior performance. Our study demonstrates the potential of graph-based models to enhance disease prediction in diverse healthcare applications.
\end{abstract}

\begin{keywords}
Graph neural networks, adaptive boosting, multi-modality, disease progression modelling, disease prediction
\end{keywords}

\section{Introduction}
\label{sec:intro}

Precision (personalized) medicine aims at providing treatments that are best for a specific patient based on individual variations in genes, environmental factors, and lifestyles. It has aroused much attention due to its potential to refine healthcare decisions for each patient. But the goal of precision medicine is hard to realize because of the vast heterogeneity in patient profiles. Thus, it is necessary to consider multi-modal data from different sources for each patient. Concurrently, machine learning (ML) techniques have demonstrated efficacy in analyzing multi-modal data \citep{tu2023towards,zhang2023biomedgpt}. Also, a growing number of large-scale, shared clinical datasets of images, genetics, and assessments \citep{hulsen2019big,shilo2020axes} accelerates the application of ML to personalized clinical prediction tasks, such as predicting disease progression or occurrence \citep{mhasawade2021machine}.

Among various ML methods, graph neural networks (GNNs) are well-suited for considering relationships between individuals, further facilitating personalized modeling and prediction. Specifically, given a dataset containing the data of many patients and each patient having multi-modal data, we can construct a graph with patients as nodes and connect similar patients, where the similarity is determined by the features of the patients. For example, patients of similar ages or same genders can be connected to form a graph \citep{parisot2017spectral}. Then, by utilizing features from the multi-modal data as node features, the GNN can be trained for predictions. This approach ensures that the modeling and prediction for a given patient are informed not only by their individual data but also by data from analogous patients. Prior studies \citep{xing2019dynamic,zheng2022multi} have underscored the efficacy of such models for population analysis and multi-modal data integration in medical fields.

However, selecting the appropriate edge features to define patient similarity is challenging since each patient has high-dimensional features from multi-modal data sources. It is also crucial because selected edge features greatly influence prediction results. Previous works rely on human expertise and prior knowledge to determine edge features. This approach lacks scalability because one needs to find new edge features for a new problem. Moreover, identifying edge features in prediction tasks related to complex diseases can be non-trivial, even for human experts.

To address the above issue, we propose a novel algorithm named \ours, which can automatically select important features to construct multiple patient similarity graphs. Generally, in \ours, GNNs as weak classifiers are iteratively trained in the adaptive boosting (AdaBoost) process. In each round, the most important feature with a certain criteria is selected as the edge feature, and an edge is established when the gap of this feature between two patients is less than a threshold, which is also determined by the algorithm. Then, a GNN is trained based on the constructed graph. Finally, all trained GNNs are ensembled for prediction. Notably, \ours{} can also be compatible with human prior knowledge by involving graphs built by human experts in the final ensemble model. Therefore, inter-individual information and intra-individual features are well-unified in one model, with human efforts on graph building largely relieved by \ours.

% In summary, \textit{AdaMedGraph} can automatically construct patient similarity graphs 

% automatic, transparency

We conduct extensive experiments on two real-world medical scenarios, i.e., Parkinson’s disease (PD) progression and metabolic syndrome prediction. \ours{} shows superior performance on almost all tasks compared with some strong baselines.

\begin{figure*}[htb]
\vspace{-3em}
\centering
 % Caption and label go in the first argument and the figure contents
 % go in the second argument
\floatconts
  {fig:Figure1a}
  %{\caption{AdaMedGraph: For every feature, a relation graph was bulit and an APPNP model was trained as weak classifier. Utilizing the ADaBoost algorithm, the most  optimal weighted weak classifiers were automatically selected. Weighted sum the  weak classifiers and get the final predictions.}}

  {\setlength{\abovecaptionskip}{0.cm}}{\includegraphics[width=1\linewidth]{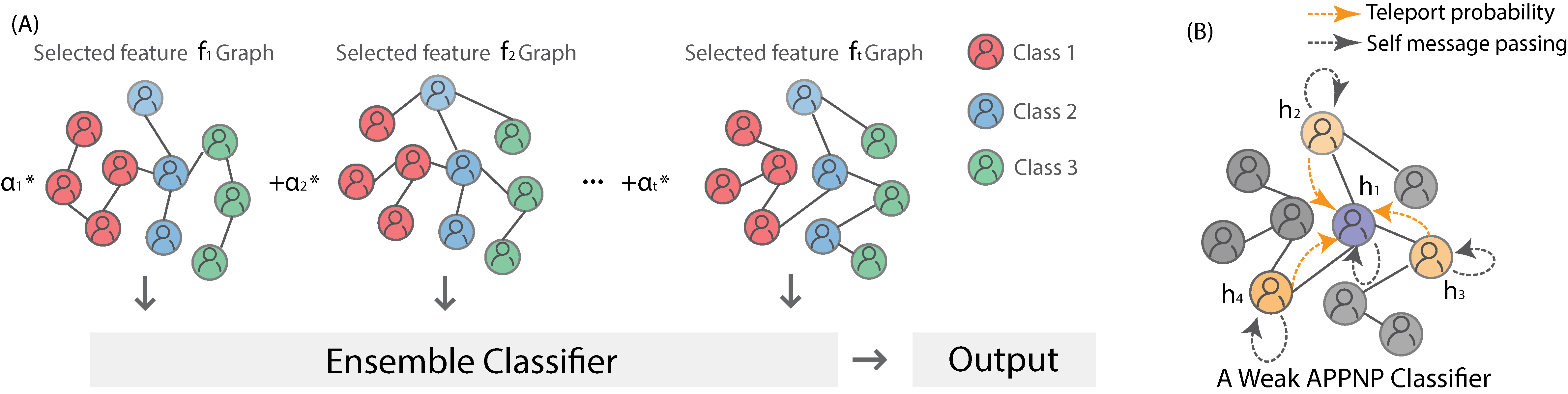}}
  {\caption{Overview of AdaMedGraph: (A) we iteratively select $T$ features to train $T$ weak classifiers (i.e., APPNP models), and combine them as an ensemble classifier to provide the final prediction.  
  (B) Illustration of features propagation within APPNP weak classifier.}}
\vspace{-1em}
\end{figure*}

% The main contributions of our paper are:
% \begin{enumerate}
%     \item We propose \ours{} for handling prediction tasks in personalized medicine. \ours{} can automatically construct similarity graph between patients and utilize multi-modal information via GNNs. Inter-individual information and intra-individual features are well-unified in one model, with human efforts on graph building largely relieved.
%     \item experiments
% \end{enumerate}

% \textbf{Contribution}
% The main contributions of our paper are:
% \begin{enumerate}
%     \item We proposed a multi-modality model leveraging a combination of inter-individual and intra-individual data, allowing for distinct prediction models tailored to each patient, which improved classification accuracy; 
%     \item  Our model consists of automated algorithms for graph construction and takes into account multiple relationships
%     \item Our model has the capacity for transparency and interpretability, enabling it to automatically create the significance of patient-relation characteristics and individual-prediction features.

% \end{enumerate}
\section{Related Work}
\label{sec:rwork}
General tabular data processing methods have been widely used to solve medical-related tasks. While these methods based on feature interactions, such as LR and Wide \& Deep, as well as gradient boosting techniques like Gradient Boosting Decision Tree (GBDT) and XGBoost, have shown promise in handling tabular medical data, they tend to focus on feature interactions and often overlook the significance of relationships between patients.

Comparing with general tabular data processing methods, GNNs have the capability to generate embeddings for individual instances by leveraging their inherent information and iteratively gathering messages from neighboring nodes \citep{gilmer2017neural}. Well-established GNN methods encompass GCN \citep{kipf2016semi}, GraphSAGE \citep{hamilton2017inductive}, graph attention networks (GAT) \citep{velickovic2017graph}, and graph isomorphism networks (GIN) \citep{xu2018powerful}. While GNNs excel at capturing intricate node relationships, they are applicable exclusively when dealing with data structured in a graph format. And the quality of the underlying graph structures significantly influences GNN performance \citep{DBLP:journals/corr/abs-2103-03036}. Given that medical tabular data lacks graph structures, the construction of meaningful graphs as inputs for GNNs becomes a critical consideration in this context.

Graph construction for tabular data typically falls into four categories: rule-based, learning-based, search-based, and knowledge-based \citep{li2023graph}.
(1) The rule-based approach operates by either leveraging inherent data dependencies among data instances and features, e.g., \citet{Cvitkovic2020, guo2021tabgnn, you2020handling, zhu2003semi} , or by relying on manually specified heuristics, e.g., \citet{goodge2022lunar, rocheteau2021predicting}. The knowledge-based approach uses domain experts to provide insights into the relationships between data instances, enabling fine-grained graph construction \citep{du2021tabularnet}. These methods require extra heuristics or knowledge. 
(2) The learning-based approach for graph construction automatically creates edges between nodes by treating the adjacency matrix as a parameter \citep{hettige2020medgraph, liu2022towards}. However, their internal structures often remain opaque and difficult to understand \citep{xia2021graph}.
%Comparing with them, \ours{} offers another advantage on interpretability, as their internal structures often remain opaque and difficult to understand \citep{xia2021graph}, while \ours{} explicitly selects edge features. 
(3) The search-based approach often involves neural architecture search to discover improved graph topologies for representation learning, as demonstrated in \citep{xie2021fives}. However, it prioritizes modeling interactions between features rather than samples or patients. Another search-based approach \citep{du2022learning} models data as a bipartite graph, where features with a specific value are regarded as nodes. Such a graph construction method generally excludes numerical variables from the model. 
%\ours{} models interactions between patients, and do not have the above constrain.
Comparing with the current methods, our \ours{} models interactions between patients without the above constraints, provides meaningful automatic graphs constructions, offers an advantage in interpretability, and improves prediction performances.

\section{Methods}
\label{sec:method}
Given $N$ patients, we have the input $X \in \mathbb{R}^{N*M}$ as patient features, and $Y \in \mathbb{R}^{N*K}$ as the one-hot encoding labels. 
$M$ is the number of features extracted from multi-model data, and $K$ is the number of label classes.
We propose \ours{} to identify different relationships among patients automatically and then classify the patients accurately.
The whole process is similar to AdaBoost, with GNNs as weak classifiers. In iteration $t$, to specify each weak classifier $g_{\theta_t}$, we need to first select a certain feature $f_t$ to measure the similarity of patients to build an adjacency matrix $A_t$, and then decide the parameters ${\theta_t}$ and $\alpha_t$ by training $g_{\theta_t}(X, A_t)$. The overall objective is to minimize the exponential loss $L(g_T(X), Y)$ where
$$
g_T(X) = \sum_{t=1}^{T} \alpha_t * g_{\theta_t}(X, A_t),
$$
by finding the optimal $\{f_t\}_{t=1,...,T}$, and $\{\theta_t\}_{t=1,...,T}$. $\{\alpha_t\}_{t=1,...,T}$ denote the weights to ensemble GNNs. \figureref{fig:Figure1a} presents the overview of \textit{AdaMedGraph}, and we will dive into the details in the following.

%Given an input graph $\mathcal{G}=\{V,E\}$, where $V$ denote the set of $N$ nodes, and $E$ describe the set of edges. The corresponding adjacency matrix is $A \in R^{N \times N}$. The objective of our methods is to automatically find the best combination of graphs ${G_i}$ where ${i=1,2,...n}$, to make accurate classification predictions for node labels. 

%\paragraph{Feature Extraction.} 

\paragraph{Initialization} 
We first standardize the format of features extracted from multi-modal data by converting them into categorical or numerical values. For example, the images (e.g., MRI) are pre-processed into different brain segments to calculate the volume of grey matter, which serves as an indicator of human cognitive abilities. Additionally, leveraging prior knowledge, genetic data is filtered to retain only the crucial genes associated with the predictions.

At the beginning of the algorithm, we assign the weight of each patient as $w_{i0} = \frac{1}{N}$ where $N$ is the total number of patients. We also specify the type of weak classifiers as Approximate Personalized Propagation of Neural Predictions (APPNP) \citep{klicpera2019predict}, a simple yet effective GNN model tailored for graph-structured data.

\paragraph{Constructing the potential $A_t$}
At iteration $t$, we have the adaptive weights $w_{i,t}$ for data of patient $i$, and the goal is to train a weak classifier $g_{\theta_t}$ and the corresponding weight $\alpha_t$ of $g_{\theta_t}$ to minimize the current weighted loss.

To construct potential adjacency matrices, two things need to be determined:
(1) the feature $f_{j,t}$, $ j \in [1, M]$ that characterizes a specific relationship among patients, and (2) the 
threshold $\gamma_t$ that determines the existence of edges in $A_t$.
%threshold $\gamma_t \in [0,1]$ that determines the existence of edges in $A_t$.  
Specifically, we explore all the features and consider the 16-quantile, 8-quantile, and 4-quantile of the selected feature values as $\gamma_t$. The edge between patient $i_1$ and $i_2$ is equal to $1$ if the absolute difference between their feature values, $|f_t(i_1) - f_t(i_2) |$, is less than or equal to $\gamma_t$; otherwise, it is set to $0$. In total, we obtain a set of $\{A_t\}$ having $3M$ elements.

% \subsection{Construction of Population Graphs}
% For a single graph, we regard every patient as a node $v$, our proposed methodology requires two design elements: firstly, the choice of the feature vector $\operatorname{x}(v)$ that characterizes each sample, and secondly, the choice of interactions between samples, accomplished through the definition of the graph edges $E$. We employ all available input information $F={f_i}$ as node features and systematically explore each input feature $f_i$ to decide the construct population graphs through an iterative process. Specifically, we define a similarity function $Sim$, and threshold $\theta_i$, to decide the connection of patient v and w using feature $f_i$:

% \begin{equation}\label{eq:eq1}
% Sim (f_i^v,f_i^w )= 
% \left\{\begin{array}{l}
%             1, \text{ if  } |f_i^v-f_i^w |< \theta_i,\\
%             0, \text{ otherwise} \\
%         \end{array}
% \right.
% \end{equation}
% In this case, we can build a set of population graph $\{G_i\}$ where each graph represents a kind of relationship between patients.

\paragraph{Training the weak classifier $g_{\theta_t}$}
Next, we aim to find an optimal choice of $A_t$, $\theta_t$ and $\alpha_t$ satisfying
$$
 \min_{A_t, \theta_t, \alpha_t}  L( g_{\theta_t}(X, A_t), Y) = \sum_{i=1}^{N} w_{i,t} \cdot \exp(- \alpha_t \cdot y_i \cdot g_{\theta_t}(x_i, A_t)).
$$
Following the SAMME \citep{zhu2006multi}, the optimal $\theta_t$ is equal to minimize the error at iteration $t$
$$ \small
\text{err}_t = \sum_{i=1}^{N} w_{i,t} \cdot \mathds{1}(y_i \neq g_{\theta_t}(x_i, A_t)),
$$
and the corresponding weight $\alpha_t$ is calculated by
$$ \small
\alpha_t = \frac{1}{2}\log \frac{1-\text{err}_t}{\text{err}_t} + \log(K-1).
$$ 
For every $A_t$, we can train an APPNP model to obtain $\theta_t$ and $\alpha_t$. Thus, we explore each possible value in the previous constructed $\{A_t\}$ to find a minimal $ L( g_{\theta_t}(X, A_t), Y)$.

After determining the optimal solution, the weights of the patients are updated by
$$
w_{i,t+1} = w_{i,t} \cdot \exp(\alpha_t \cdot \mathds{1}(y_i \neq g_{\theta_t}(x_i, A_t) ),
$$
and we proceed to the next iteration.
\paragraph{Termination}
The whole process is ended at the iteration $T$ when the error rate of $g_T(X)$ or $g_{\theta_T}(X, A_T)$ gets equal or larger than $\frac{K-1}{K}$.

% \subsection{AdaBoost with Graph Neural Network as weak classifier}
% The AdaMedGraph framework leverages the power of Graph Neural Network (GNN) \citep{scarselli2008graph} by aggregating them using the AdaBoost \citep{hastie2009multi} algorithm to enhance predictive performance. Specifically, the AdaMedGraph classifier framework can be write as:
% \begin{equation}\label{eq:eq2}
% F(x(v))= \sum_{i=1}^{M}\alpha_i*g_i (x(v),G_i)
% \end{equation}
% where $F(x(v))$is the ensemble classifier obtained after m rounds of training, and $x(v)$ denotes node features. For each round, the framework automatically adds a trained GNN classifier $g_i (x(v),G_i)$ to improve the prediction performance. The weight of the classifier $\alpha_i$ denotes the importance of classifier (also the kind of relation between patients), and it could be obtained according to the error of the classifier during training. See \figureref{fig:Figure1a} for the overview of our proposed model.

\begin{table*}[]
\vspace{-4em}
\caption{PPMI and PDBP 24-month prediction AUROC score. M* is MDS-UPDRS.}
\label{tab:PPMIPDBP}
\scalebox{0.8}{ 
\begin{tabular}{lllllllllllll}
\hline
                                                           & \multicolumn{2}{c}{\textbf{MLP}} & \multicolumn{2}{c}{\textbf{LR}} & \multicolumn{2}{c}{\textbf{SVM}} & \multicolumn{2}{c}{\textbf{RF}} & \multicolumn{2}{c}{\textbf{XGB}} & \multicolumn{2}{c}{\textbf{AdaMedGraph}} \\ \hline
Label                                                      & PPMI            & PDBP           & PPMI           & PDBP           & PPMI            & PDBP           & PPMI           & PDBP           & PPMI            & PDBP           & PPMI              & PDBP              \\ \hline
HY                                                         & 0.593           & 0.450          & 0.652          & 0.504          & 0.705           & 0.614          & 0.664          & 0.692          & 0.717           & 0.643          & \textbf{0.775}    & \textbf{0.701}    \\ \hline
\begin{tabular}[c]{@{}l@{}}M* I\end{tabular}      & 0.445           & 0.473          & 0.601          & 0.541          & 0.573           & 0.554          & 0.605          & 0.581          & 0.625           & 0.584          & \textbf{0.656}    & \textbf{0.586}    \\ \hline
\begin{tabular}[c]{@{}l@{}}M* II\end{tabular}     & 0.441           & 0.499          & 0.578          & 0.536          & 0.570           & 0.610          & 0.602          & 0.598          & 0.620           & 0.602          & \textbf{0.651}    & \textbf{0.621}    \\ \hline
\begin{tabular}[c]{@{}l@{}}M* III\end{tabular}    & 0.695           & 0.645          & 0.691          & 0.648          & 0.695           & 0.645          & 0.670          & 0.643          & 0.710           & 0.610          & \textbf{0.717}    & \textbf{0.660}    \\ \hline
\begin{tabular}[c]{@{}l@{}}M* Total\end{tabular}  & 0.615           & 0.577          & 0.635          & 0.587          & 0.634           & 0.598          & 0.595          & 0.579          & 0.613           & 0.590          & \textbf{0.646}    & \textbf{0.601}    \\ \hline
\begin{tabular}[c]{@{}l@{}}M* Axial\end{tabular}  & 0.618           & 0.552          & 0.637          & 0.653          & 0.661           & 0.652          & 0.645          & 0.657          & 0.627           & 0.618          & \textbf{0.661}    & \textbf{0.660}    \\ \hline
\begin{tabular}[c]{@{}l@{}}M* Rigid\end{tabular}  & 0.644           & 0.659          & 0.665          & 0.656          & 0.654           & 0.678          & 0.663          & 0.658          & 0.591           & 0.626          & \textbf{0.694}    & \textbf{0.683}    \\ \hline
\begin{tabular}[c]{@{}l@{}}M* Tremor\end{tabular} & 0.669           & 0.651          & 0.689          & 0.635          & 0.684           & 0.619          & 0.713          & 0.625          & 0.672           & 0.613          & \textbf{0.715}    & \textbf{0.685}    \\ \hline
MoCA                                                       & 0.519           & 0.497          & 0.600 & \textbf{0.694} & 0.574           & 0.559          & 0.573          & 0.623          & 0.614           & 0.684          & \textbf{0.660}             & 0.506             \\ \hline
ESS                                                        & 0.573           & 0.546          & 0.573          & 0.630          & 0.601           & 0.628          & 0.625          & 0.625          & 0.611           & 0.628          & \textbf{0.645}    & \textbf{0.631}    \\ \hline
\end{tabular}
}
\end{table*}

\section{Data and Experiments}
\label{sec:experiments}

% We implemented PD progression and metabolic syndrome incidence prediction experiments.

In this section, we first introduce prediction tasks and settings in two real-world medical scenarios and then present the results of these tasks.

\subsection{PD progression}
The longitudinal data from the Parkinson's Progression Markers Initiative (PPMI) \citep{marek2018parkinson} and the Parkinson's Disease Biomarkers Program (PDBP) \citep{gwinn2017parkinson}are used in our work. 

\paragraph{Prediction tasks}
We select 10 assessment scores as labels for disease progression: Movement Disorder Society-sponsored Revision of the Unified Parkinson's Disease Rating Scale (MDS-UPDRS) I, II, III, and total (sum of I, II, and III), and the sub-scores for tremors, axial symptoms, and rigidity in MDS-UPDRS III; Hoehn and Yahr Scale (HY); Montreal Cognitive Assessment (MoCA) total; and Epworth Sleepiness Scale (ESS) total. To get the classification labels, we compute the mean and standard deviation of changes in each score (except HY since HY has a standard classification scale) from baseline to 24-months in the healthy control (HC) group and compare them with the changes in the PD group. By constructing a reference interval based on the HC changes, we then categorize the PD changes as ``No Change" if within the interval, or as ``Better" or ``Worse" if less or more than the interval, respectively.

\paragraph{Feature extraction}
We pre-possess genetic data and magnetic resonance imaging (MRI) before training. For MRI, the participant's T1 scan has been processed into 138 segments, and the volumes of the segments are used as input. For genetic data, the monogenic mutation statuses of GBA, LRRK2, SNCA, and APOE\_E4 are encoded as binary variables.

\paragraph{Dataset statistics and division} 
Two sets of experiments have been designed for the PD progression task. In PD Experiment I, we utilize clinical assessments and genetic data as input features, comprising a total of 78 variables, given the absence of MRI data in the PDBP dataset. The PPMI dataset is split into a training set (80\%, n = 358) and an internal testing set (20\%, n = 90). Besides, we randomly select 20\% of the training data as our validation set during training to select the final best models. The PDBP dataset serves as an external validation dataset (n = 228). 

In Experiment II, we expand the range of input features to include clinical assessments, MRI data, and genetic information, resulting in a total of 234 variables. This study exclusively utilizes the PPMI dataset, which has been divided into an 80\% training set (n = 252) and a 20\% testing set (n = 65). Similarly, 20\% of training data has been randomly separated as validation set. In both experiments, the train-test split is based on participants' enrollment times, with the earlier visit participants assigned to the training set while the later visit participants are allocated to the testing set, while the validation split is done randomly.

\paragraph{Baselines}
To evaluate the performance of our model, we include several ML classification methods as baseline models. The aforementioned models have frequently been employed in medical classification tasks, encompassing a 2-layer Multilayer Perceptron (MLP), a logistic regression classifier (LR), a random forest classifier (RF), a support vector machine classifier (SVM), and a XGBoost model.
Furthermore, in order to comprehend the importance of automatically selecting edge features, we construct a graph based on prior knowledge. Since it is well-known that age is a critical factor for PD progression, we construct a graph based on age (with a threshold of 5) and develop an APPNP (referred to as APPNP-Age) as a comparative model in experiment II. 

\subsection{Metabolic syndrome}
In the metabolic syndrome prediction task, we include 109,027 individuals in the UK Biobank \citep{sudlow2015uk} (80\% for training and 20\% for testing). All subjects have no record of metabolic syndrome on their baseline screen, and their 168 proton nuclear magnetic resonance (NMR) metabolic biomarkers and 17 traditional clinical risk variables are used as input \citep{lian2023metabolic} to predict whether participants will have the diseases after a period of time. The input data has been processed into 9 principal components using principal component analysis, and XGBoost and TabNet \citep{arik2021tabnet} are implemented as baseline models.
% to compare with \ours{}.

\subsection{Results}
\label{sec:result}

% We utilize the weighted area under the receiver operating characteristic curve (AUROC) for multiclass task \citep{bishop2006pattern} as the evaluation metric for PD task and AUROC for Metabolic syndrome.

We utilize the weighted area under the receiver operating characteristic curve (AUROC) as the evaluation metric for all tasks.

\paragraph{PD progression}
In the context of predicting 24-month progression experiment I, our AdaMedGraph model has received greater accuracy in comparison to all baseline models across all labels, with the exception of MoCA in the PDBP dataset. This discrepancy can be attributed to differences in the distribution of selected edge features for MoCA between the two datasets. These superior results from our model underscore the significance of incorporating both intra- and inter-patient data for individual disease progression prediction. Refer to the \tableref{tab:PPMIPDBP} for a comprehensive understanding of the performances.
% In terms of 24-month HY prediction, AdaMedGraph outperformed XGBoost, RF, SVM, LR, and MLP with AUC values of 0.775 (PPMI) and 0.701 (PDBP), compared to the AUC values of 0.717 and 0.643 for XGBoost, 0.664 and 0.692 for RF, 0.705 and 0.614 for SVM, 0.652 and 0.504 for LR, and 0.593 and 0.450 for MLP. 

In experiment II, AdaMedGraph also exhibits superior performance when compared to all the baseline models as well as APPNP-age, which suggests that taking into account automatically selected relationships among patients can significantly enhance the performance of prediction. Please refer to \tableref{tab:PPMI} for details.

\begin{table}[h]
\caption{PPMI 24-month prediction AUROC score. A-A is APPNP-Age. M* is MDS-UPDRS.}
\label{tab:PPMI}
\scalebox{0.75}{
\begin{tabular}{llllllll}
\hline
Label                                                      & MLP   & LR    & SVM   & RF    & XGB            & A-A & Ours       \\ \hline
HY                                                         & 0.556 & 0.630 & 0.594 & 0.633 & 0.643          & 0.630 & \textbf{0.682} \\ \hline
\begin{tabular}[c]{@{}l@{}}M* I\end{tabular}      & 0.543 & 0.638 & 0.531 & 0.593 & 0.628          & 0.699 & \textbf{0.730} \\ \hline
\begin{tabular}[c]{@{}l@{}}M* II\end{tabular}     & 0.507 & 0.588 & 0.573 & 0.457 & 0.575          & 0.621 & \textbf{0.652} \\ \hline
\begin{tabular}[c]{@{}l@{}}M* III\end{tabular}    & 0.409 & 0.583 & 0.587 & 0.553 & 0.566          & 0.604 & \textbf{0.666} \\ \hline
\begin{tabular}[c]{@{}l@{}}M* Total\end{tabular}  & 0.648 & 0.617 & 0.593 & 0.590 & 0.598          & 0.563 & \textbf{0.684} \\ \hline
\begin{tabular}[c]{@{}l@{}}M* Axial\end{tabular}  & 0.517 & 0.610 & 0.610 & 0.549 & 0.538          & 0.591 & \textbf{0.689} \\ \hline
\begin{tabular}[c]{@{}l@{}}M* Rigid\end{tabular}  & 0.419 & 0.572 & 0.636 & 0.548 & 0.628          & 0.682 & \textbf{0.693} \\ \hline
\begin{tabular}[c]{@{}l@{}}M* Tremor\end{tabular} & 0.520 & 0.628 & 0.648 & 0.657 & \textbf{0.700} & 0.585 & \textbf{0.700} \\ \hline
MoCA                                                       & 0.680 & 0.713 & 0.672 & 0.714 & 0.723          & 0.668 & \textbf{0.746} \\ \hline
ESS                                                        & 0.537 & 0.623 & 0.639 & 0.645 & 0.596          & 0.649 & \textbf{0.676} \\ \hline
\end{tabular}}
\end{table}
\vspace{-10pt} 

\paragraph{Metabolic syndrome}
In the context of predicting the occurrence of metabolic syndrome, the AdaMedGraph model demonstrates superior performance with an area under the AUROC of 0.675 on the testing dataset. This surpasses both the XGBoost model, which achieved an AUROC of 0.641, and the TabNet model, which achieved an AUROC of 0.672.

\section{Discussion}
\label{sec:discussion}
In this study, we introduce an innovative algorithm, namely \ours{}, designed to autonomously identify important features for the construction of multiple patient similarity graphs, which serve as the basis for training GNNs in an AdaBoost framework, thereby enhancing the accuracy of classification tasks. We have conducted two sets of clinical experiments, and the results affirm our initial hypothesis that automatically constructing multi-relationship graphs among patients using inter- and intra-individual data can benefit personalized medicine.

One thing that needs to be noticed is the computational cost of our \ours{}. Although our model goes through all potential features to construct a total of $3M$ graphs during training, it's important to note that we use the APPNP model as the weak classifier, known for its relatively low computational cost, with only linear computational complexity. In this case, the total computational cost of our model is acceptable. In our experiments, we have observed reasonable training time costs. For instance, in our PD prediction task with 234 input features and 317 patient samples, the training process has taken less than 5 minutes. Similarly, for the metabolic syndrome prediction task, which includes 109,027 patients and 9 features, it takes less than 30 minutes for training. These experiments have been conducted on a single V100 GPU. Furthermore, to optimize for high-dimensional settings, a simple feature selection process could be integrated before searching for the best features.

% \begin{table}[htb]
% \centering
% \setlength{\tabcolsep}{14mm}
% {
% \begin{tabular}{@{}ll@{}}
% \toprule
% Model    & AUC \\ \midrule
% XGBoost  & 0.641  \\
% TabNet   & 0.671  \\
% AdaMedGraph & 0.675  \\ \bottomrule
% \end{tabular}
% }
% \caption{Metabolic syndrome prediction performance of XGBoost, TabNet and AdaMedGraph}
% \label{tab:mate}
% \end{table}

% \acks{}

\bibliography{jmlr-sample}

\newpage

\appendix
\section{Ethics Statement}\label{apd:first}
The population cohort in the metabolic syndrome experiments is from the UK Biobank [Application Number 78730] which has received ethical approval from the North West Multicentre Research Ethics Committee (REC reference: 11/NW/03820). All participants have given written informed consent before enrolment.

PD study data is retrieved from AMPPD platform https://amp-pd.org/ using the terms of the AMP-PD Data Use Agreement and PPMI website https://www.ppmi-info.org/. The data has used in this study has been collected before the commencement of our study and has been obtained in an anonymized format. In the original PPMI and PDBP studies, participants have previously consented in writing to the sharing of their data. These investigations have been conducted in accordance with protocols authorized by the Indiana University Institutional Review Board (IRB) for PPMI and the respective PDBP centers.
% This is the first appendix.
\section{Hyper-parameters}\label{apd:first}
We list the hyper-parameters of baseline models and \ours{} in this part. Grid search method has used for all models.
\subsection{Task 1}
\paragraph{MLP} We train the MLP models with Adam optimizer and search the hyper-parameters of hidden dimension in \{128, 256, 512, 1024\},  L2 regularization in \{$10^{-5}$, $10^{-4}$, $10^{-3}$\}, learning rate \{$10^{-5}$, $5*10^{-4}$, $10^{-4}$, $5*10^{-3}$\}, dropout in \{0.1, 0.3, 0.5\} and total epoch in \{100, 200, 300\}.

\paragraph{LR} We search the LR models hyper-parameters of regularization method in \{L1, L2, elastic net\} and strength of the regularization in \{0.01, 0.1, 1, 10\}.

\paragraph{SVM} We search the the LR models hyper-parameters of kernels in \{linear, polynomial, radial basis function\} and strength of the regularization in in \{0.01, 0.1, 1, 10\}.

\paragraph{RF} We search the LR models hyper-parameters of number of estimators in \{50, 100, 150, 200 \}, max depth in \{10, 20, 30, 40, 50\}, minimum number of samples required to be at a leaf node in \{1, 2, 4\}, and the minimum number of samples required to split an internal node in \{1, 2, 3\}.

\paragraph{XGB} We list  the XGB models hyper-parameters searching in  \tableref{tab:xgbtable}.

\begin{table}[h]
\caption{XGBoost hyper-parameters search list}
\label{tab:xgbtable}
\begin{tabular}{|l|l|lll}
\cline{1-2}
Hyper-parameter      & Tunning Range               &  &  &  \\ \cline{1-2}
Number of estimators & \{50, 75, 100, 150, 200\}   &  &  &  \\ \cline{1-2}
Max depth            & \{1, 2, 3, 6, 9\}           &  &  &  \\ \cline{1-2}
Learning rate     & \{0.001, 0.01, 0.1, 0.3, 0.5\}           &  &  &  \\ \cline{1-2}
Min child weight     & \{1, 2, 3, 4, 5\}           &  &  &  \\ \cline{1-2}
Gamma                & \{ 0.5,  1,  2, 3, 6, 9\}   &  &  &  \\ \cline{1-2}
Colsample by tree    & \{0, 0.25, 0.5, 0.75, 1\}   &  &  &  \\ \cline{1-2}
Colsample by level   & \{0, 0.25, 0.5, 0.75, 1\}   &  &  &  \\ \cline{1-2}
lambda               & \{0.05, 0.1, 0.25, 0.5, 1\} &  &  &  \\ \cline{1-2}
alpha                & \{0.5, 1, 10\}              &  &  &  \\ \cline{1-2}
\end{tabular}
\end{table}

\paragraph{APPNP-Age} We train our APPNP-Age with Adam optimizer, Cosine Annealing Scheduler for weight decay, and 100 total epochs. We search the hyper-parameters of hidden dimension of linear layer (H) in \{128, 256, 512\}, number of iterations (k) in \{3, 5, 10\}, teleport probability in \{0.1, 0.3, 0.5, 1\}, dropout in \{0.1, 0.3, 0.5\}, learning rate in \{$10^{-5}$, $5*10^{-4}$, $10^{-4}$, $5*10^{-3}$ \}, and threshold for age in \{4, 4.5, 5, 5.5, 6\}.

\paragraph{AdaMedGraph} The training details for AdaMedGraph contain two parts: training a single APPNP and training AdaBosst. The details have been summarized in \tableref{tab:adamed}.

\begin{table}[h]
\caption{AdaMedGraph hyper-parameter list. Hyper-parameters in braces are searched.}
\label{tab:adamed}
\scalebox{0.95}{
\begin{tabular}{|ll|}
\hline
\multicolumn{2}{|c|}{Single APPNP}                                                   \\ \hline
\multicolumn{1}{|l|}{Optimizer}                        & Adam                        \\ \hline
\multicolumn{1}{|l|}{Weight decay}                     & \{0.00001,  0.0001, 0.001\} \\ \hline
\multicolumn{1}{|l|}{Learning rate decay}              & Cosine Annealing Scheduler  \\ \hline
% \multicolumn{1}{|l|}{Learning rate}                    & \{$10^{-5}$,  $5*10^{-5}$, $10^{-4}$,  $5*10^{-4}$,  $10^{-3}$, $5*10^{-3}$\} \\ \hline
\multicolumn{1}{|l|}{Learning rate}                    & \begin{tabular}[c]{@{}l@{}}\{$10^{-5}$,  $5*10^{-5}$, $10^{-4}$,  \\ $5*10^{-4}$,  $10^{-3}$, $5*10^{-3}$\}\end{tabular} \\ \hline
\multicolumn{1}{|l|}{Total epoch}                      & \{100, 150, 200\}           \\ \hline
\multicolumn{1}{|l|}{Early stop}                       & \{5, 10, 20\}               \\ \hline
\multicolumn{1}{|l|}{H} & \{128, 256, 512\}           \\ \hline
\multicolumn{1}{|l|}{k}                                & \{3, 5, 10\}                \\ \hline
\multicolumn{1}{|l|}{Teleport probability}             & \{0.1, 0.3, 0.5, 1\}        \\ \hline
\multicolumn{1}{|l|}{Dropout}                          & \{0.1, 0,3, 0,5\}           \\ \hline
\multicolumn{2}{|c|}{AdaBoost}                                                       \\ \hline
\multicolumn{1}{|l|}{Number of estimators}              & \{5, 10, 15, 20\}           \\ \hline
\multicolumn{1}{|l|}{Learning rate}                    & \{0.1, 0.33, 0.5, 0.66, 1\}       \\ \hline
\end{tabular}}
\end{table}

\subsection{Task 2: metabolic syndrome prediction}

\paragraph{XGB} We list  the XGB models hyper-parameters searching in  \tableref{tab:xgbtable2}.
\vspace{-2em}
\begin{table}[h]
\caption{XGBoost hyper-parameters search list for metabolic syndrome prediction}
\label{tab:xgbtable2} 
\begin{tabular}{|l|l|lll}
\cline{1-2}
Hyper-parameter      & Tunning Range               &  &  &  \\ \cline{1-2}
Number of estimators & \{50, 100, 200, 300\}   &  &  &  \\ \cline{1-2}
Max depth            & \{3, 6, 9, 12\}           &  &  &  \\ \cline{1-2}
Learning rate     & \{0.001, 0.01, 0.1, 0.3, 0.5\}           &  &  &  \\ \cline{1-2}
Min child weight     & \{1, 2, 3, 4, 5\}           &  &  &  \\ \cline{1-2}
Gamma                & \{ 0.5,  1, 3, 6, 9, 12\}   &  &  &  \\ \cline{1-2}
Colsample by tree    & \{0, 0.5, 1\}   &  &  &  \\ \cline{1-2}
Colsample by level   & \{0, 0.5, 1\}   &  &  &  \\ \cline{1-2}
lambda               & \{0.05, 0.1, 0.25, 0.5, 1\} &  &  &  \\ \cline{1-2}
alpha                & \{0.001, 0.01, 0.5, 1, 10\}              &  &  &  \\ \cline{1-2}
\end{tabular}
\end{table}

\paragraph{TabNet} We Search the TabNet models hyper-parameters of mask type in \{”entmax”, ”sparsemax”\}, the width of the decision prediction layer and the attention embedding for each mask in \{32, 48, 56, 64\}, number of steps in \{1, 2, 3\}, gamma in \{1.0, 1.2, 1.4\}, the number of shared Gated Linear Units at each step in \{1, 2, 3\}, learning rate in \{$10^{-5}$,  $5*10^{-5}$, $10^{-4}$, $5*10^{-4}$,  $10^{-3}$, $5*10^{-3}$\} and lambda space in \{$10^{-5}$,  $10^{-4}$\}.

\paragraph{AdaMedGraph} The training details for AdaMedGraph on task 2 have been summarized in \tableref{tab:adamed2}.
\begin{table}[!h]
\vspace{-2em}
\caption{AdaMedGraph hyper-parameter for metabolic syndrome prediction. Hyper-parameters in braces are searched.}
\label{tab:adamed2}
\scalebox{0.95}{
\begin{tabular}{|ll|}
\hline
\multicolumn{2}{|c|}{Single APPNP}                                                   \\ \hline
\multicolumn{1}{|l|}{Optimizer}                        & Adam                        \\ \hline
\multicolumn{1}{|l|}{Weight decay}                     & \{0.00001,  0.0001, 0.001\} \\ \hline
\multicolumn{1}{|l|}{Learning rate decay}              & Cosine Annealing Scheduler  \\ \hline
% \multicolumn{1}{|l|}{Learning rate}                    & \{$10^{-5}$,  $5*10^{-5}$, $10^{-4}$,  $5*10^{-4}$,  $10^{-3}$, $5*10^{-3}$\} \\ \hline
\multicolumn{1}{|l|}{Learning rate}                    & \begin{tabular}[c]{@{}l@{}}\{$10^{-5}$,  $5*10^{-5}$, $10^{-4}$,  \\ $5*10^{-4}$,  $10^{-3}$, $5*10^{-3}$\}\end{tabular} \\ \hline
\multicolumn{1}{|l|}{Total epoch}                      & \{100, 200, 500, 1000\}           \\ \hline
\multicolumn{1}{|l|}{Early stop}                       & \{10, 20, 30\}               \\ \hline
\multicolumn{1}{|l|}{H} & \{128, 256, 512\}           \\ \hline
\multicolumn{1}{|l|}{k}                                & \{3, 5, 10\}                \\ \hline
\multicolumn{1}{|l|}{Teleport probability}             & \{0.1, 0.3, 0.5, 1\}        \\ \hline
\multicolumn{1}{|l|}{Dropout}                          & \{0.1, 0,3, 0,5\}           \\ \hline
\multicolumn{2}{|c|}{AdaBoost}                                                       \\ \hline
\multicolumn{1}{|l|}{Number of estimators}              & \{5, 10, 15, 20\}           \\ \hline
\multicolumn{1}{|l|}{Learning rate}                    & \{0.1, 0.33, 0.5, 0.66, 1\}       \\ \hline
\end{tabular}}
\end{table}

% This is the second appendix.

\end{document}